\documentclass{article}

 \usepackage[preprint]{neurips_2026}

\usepackage[utf8]{inputenc} 
\usepackage[T1]{fontenc}    
\usepackage{hyperref}       
\usepackage{url}            
\usepackage{booktabs}       
\usepackage{amsfonts}       
\usepackage{nicefrac}       
\usepackage{microtype}      
\usepackage{xcolor}         
\usepackage{graphicx}
\usepackage{multirow}
\usepackage{algorithm}
\usepackage{algorithmic}
\usepackage{amsmath}
\usepackage{wrapfig}
\usepackage{caption}

\title{Rethinking Reservoir Pruning: A Dynamical Perspective for Echo State
Networks}

\author{
  Sudip Laudari\thanks{Independent Researcher.} \\
  \texttt{sudip.laudari@sydney.edu.au}
  \and
  Puspa Raj Adhikari\thanks{Advanced Materials Science and Engineering, Sungkyunkwan University.} \\
  \texttt{info11pusp@skku.edu}
}

\begin{document}

\maketitle


\begin{abstract}
Echo State Networks (ESNs) offer an efficient framework for temporal prediction, but their randomly initialized reservoirs are often over-parameterized and dynamically redundant. Existing pruning methods largely rely on static connectivity or activation statistics, which may overlook neurons that shape input-driven state transitions. We propose Dynamical Mode Pruning (DMP), a reservoir pruning method that ranks neurons by their contribution to dominant transition modes obtained from a trajectory-averaged Jacobian Gramian. DMP removes low-impact units and retrains only the readout. Experiments on chaotic and real-world time-series benchmarks show that DMP improves or preserves forecasting accuracy while reducing redundant reservoir components. Our results suggest that dynamical influence is a useful criterion for reservoir refinement beyond static structural importance alone.
\end{abstract}



\section{Introduction}
\label{introduction}

Echo State Networks (ESNs) represent a well-established paradigm in
reservoir computing, characterized by a randomly initialized recurrent layer in which only the linear readout is trained
\citep{jaeger2001echo,lukovsevivcius2009reservoir}. By avoiding backpropagation through time, this design remains computationally efficient while achieving strong performance across diverse temporal prediction tasks \citep{verstraeten2007experimental,ma2019convolutional}.

Despite these advantages, designing an effective reservoir remains challenging. Critical parameters such as spectral radius, leaky rate, input scaling, and reservoir size are commonly selected through heuristic or computationally expensive searches \citep{jaeger2007optimization}. Practitioners therefore often employ large reservoirs to ensure sufficient
expressive capacity. However, such over-parameterized systems can contain substantial redundancy, increasing memory and inference cost while potentially degrading generalization and predictive stability \citep{scardapane2014effective,goswami2024feature,rodan2010minimum}. Reducing this complexity without sacrificing predictive power has therefore become an important research objective. 


Pruning offers a particularly natural means of removing this redundancy. ESNs are especially suitable for structured neuron pruning because their recurrent weights remain fixed and do not require expensive retraining after simplification \citep{liu2022broad}. Existing reservoir-pruning approaches commonly rely on weight magnitude, activation variance, or related activity statistics \citep{dutoit2009pruning,scardapane2014effective}. Although efficient, these criteria may overlook neurons whose importance emerges through temporal state transitions rather than static activity
\citep{shunshi2019one,bianchi2016investigating,
chatzikonstantinou2021recurrent}. Beyond such simple heuristics, structural methods rank neurons according to graph connectivity and centrality
\citep{bloch2023centrality,cavallaro2024sensitivity,
freeman1979centrality,latora2001efficient}. While these formulations provide a principled description of reservoir topology, they still treat the reservoir largely as a fixed graph \citep{freeman1979centrality}, ignoring how the observed input trajectory shapes recurrent state transitions over time
\citep{gallicchio2018chasing,bianchi2016investigating}. Thus, both static activity criteria and structural graph measures fail to capture the input-driven dynamical structure that defines ESN behavior.

This gap reflects a broader challenge in neural-network pruning. Classical magnitude pruning \citep{han2015learning} and the lottery-ticket hypothesis \citep{frankle2019lottery} showed that large networks often contain effective sparse subnetworks. However, advanced importance measures do not always outperform simple baselines \citep{blalock2020state,nowak2023fantastic}, and recent theory suggests that effective pruning may require information from observed data \citep{kumar2024nofree}. These findings motivate a data-dependent criterion based on each reservoir neuron's contribution to the input-driven dynamics.


\begin{wrapfigure}{r}{0.52\linewidth}
\vspace{-20pt}
\centering
\includegraphics[width=\linewidth]{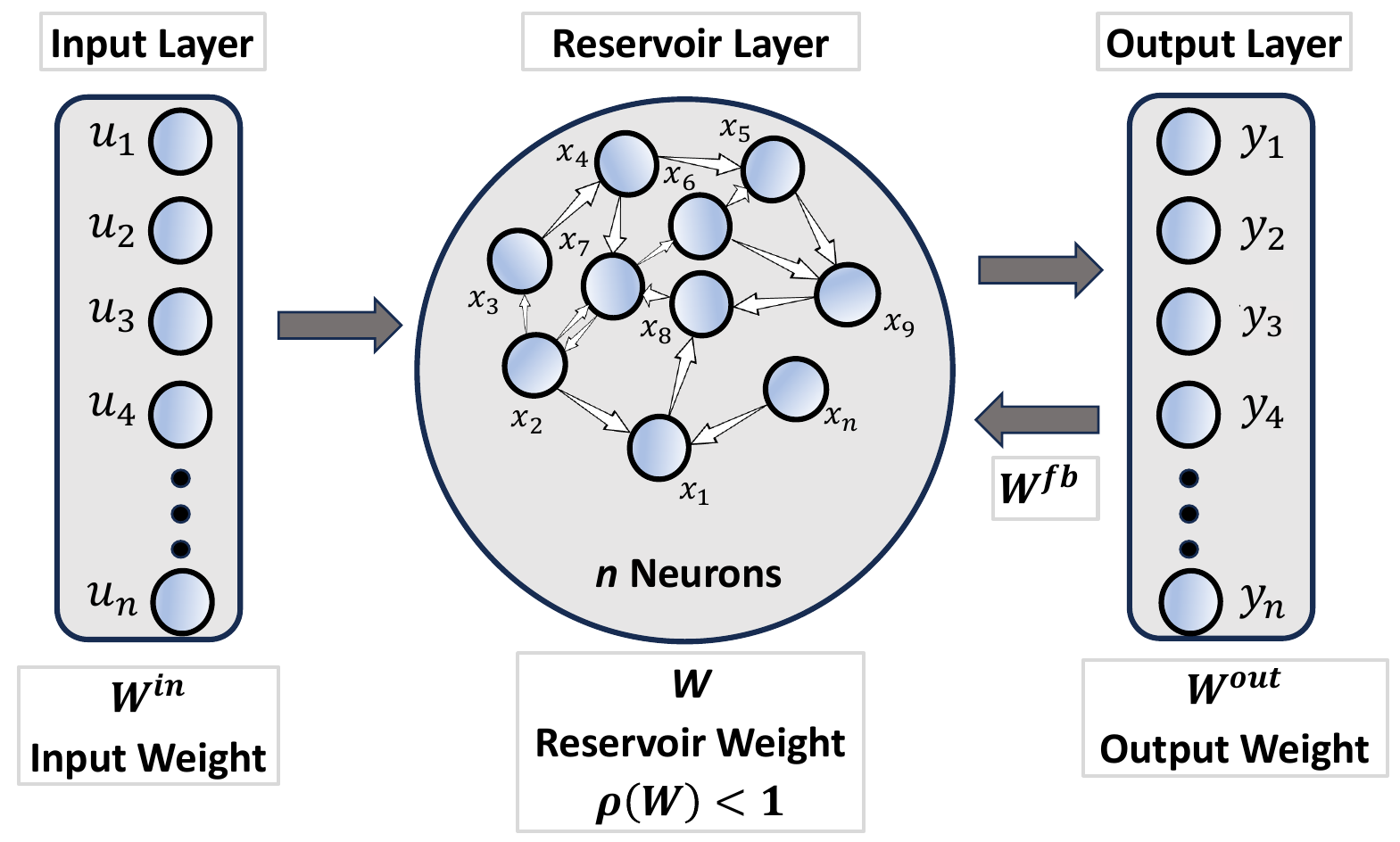}
\caption{
ESN architecture with input weights $W^{in}$, recurrent reservoir $W$, and readout $W^{out}$. Optional feedback $W^{fb}$ enables output-driven
dynamics. The reservoir is initialized to support stable input-driven state evolution.
}
\label{fig1:esn_architecture}
\vspace{-20pt}
\end{wrapfigure}



In this work, we introduce \emph{Dynamical Mode Pruning} (DMP), a one-shot structured pruning method that measures neuron importance through trajectory-dependent reservoir dynamics. DMP constructs a
trajectory-averaged Jacobian Gramian and ranks neurons according to their contribution to its dominant modes. Neurons with limited dynamical contribution are removed, after which only the linear readout is refitted. Rather than assuming that a dynamical score must always outperform simpler criteria, our objective is to provide a principled mechanism for reducing an over-parameterized ESN while retaining its dominant input-driven transition structure.


\section{Related Work}
\label{related_work}



\paragraph{Neural Network Pruning} 
Pruning has been widely studied as a strategy for reducing redundancy while preserving predictive performance. Magnitude pruning established the standard train--prune--retrain framework \citep{han2015learning}, while \citet{frankle2019lottery} showed that large randomly initialized networks may contain sparse subnetworks capable of matching the full model. Nevertheless, \citet{blalock2020state} emphasized that pruning methods must be evaluated against matched random masks, magnitude baselines, and newly initialized smaller networks. Recent studies have expanded this perspective: \citet{nowak2023fantastic} found that several pruning criteria perform similarly under controlled comparisons, with simple magnitude-based selection remaining competitive, and \citet{kumar2024nofree} further showed that effective sparse masks may require data-dependent information rather than relying only on initialization. Other work has introduced topology-aware selection \citep{yu2022topology}, submodular structured pruning \citep{elhalabi2022dataefficient}, and Bayesian model reduction \citep{wright2024bmrs}. Because parameter count alone does not guarantee practical acceleration, inference-aware approaches explicitly optimize or measure runtime speedup \citep{frantar2022spdy,kurtic2023ziplm}. Although these advances have primarily targeted deep networks with trained parameters, reservoir computing presents a distinct setting in which the recurrent reservoir remains fixed after initialization, calling for pruning criteria tailored to untrained dynamical systems.


\paragraph{Heuristic and Activation-based Reservoir Pruning}
Reservoir pruning is particularly attractive because the recurrent connections remain fixed, allowing simplification without recurrent-weight retraining \citep{liu2022broad,huang2023semi}. Early methods primarily relied on weight magnitude, node dropout, activation level, or activation variance \citep{dutoit2009pruning,scardapane2014effective}. Such criteria are computationally inexpensive, but they provide limited information about how a neuron influences recurrent state evolution. Their effectiveness may also vary non-monotonically with the pruning ratio, consistent with broader observations that increasing sparsity can alternately reduce or aggravate overfitting \citep{he2022sparse}. Consequently, there has been growing interest in more principled alternatives that account for reservoir topology and its dynamical behavior.

\paragraph{Structural and Centrality-based Methods}
Structural approaches represent the reservoir as a graph and rank neurons using degree, betweenness, closeness, eigenvector centrality, or related connectivity measures \citep{freeman1979centrality,latora2001efficient}. These ideas have been applied directly to reservoir pruning \citep{bloch2023centrality,laudari2026centrality} and extended to deep, modular, and hardware-oriented ESN architectures \citep{sun2024deep,goswami2024feature,nishioka2024high}. Other methods combine structural sensitivity with activation information to preserve prediction performance or short-term memory \citep{cavallaro2024sensitivity,wu2021chain}. While these formulations improve interpretability, they remain largely static and may not capture the evolving role of neurons in input-driven recurrent systems. In particular, weakly connected neurons can still influence dominant transition paths through their position along the observed trajectory \citep{soltani2023echo,zhu2021information}.

Complementing these topological perspectives, recent reservoir studies indicate that larger random reservoirs are not necessarily better. \citet{yadav2025task} introduced task-specific node pruning and demonstrated that performance-guided removal can reveal useful reservoir subnetworks. Similarly, compact deterministic reservoirs can compete with or outperform larger random ESNs on chaotic prediction problems \citep{martinuzzi2025minimal}. These results motivate the fresh smaller ESN and matched random-pruning baselines in our experiments. They also suggest that improvements following pruning may arise partly from removing over-parameterization rather than exclusively from the chosen importance score. Taken together, these findings underscore the need for importance criteria that go beyond static connectivity to identify which neurons actively shape the reservoir's input-driven dynamics.

\paragraph{Dynamics-Aware Model Reduction}
Existing ESN pruning methods therefore predominantly characterize static structure, activity, or limited dynamical heuristics \citep{scardapane2014effective,chatzikonstantinou2021recurrent}. Jacobian-spectrum information has previously been used to analyze and prune trained recurrent neural networks \citep{shunshi2019one}. More recently, \citet{gwak2024layer} introduced energy-based model state pruning for deep state-space models, illustrating how dominant dynamical modes can guide structured model reduction.

DMP differs from these approaches in both setting and construction. It is designed for a fixed random ESN reservoir whose recurrent weights are not trained, and it ranks neurons using a trajectory-averaged Jacobian Gramian computed from driven reservoir states. The resulting score measures each neuron's contribution to dominant input-dependent transition modes rather than relying only on connectivity, activation magnitude, or individual Jacobian entries.

Although this perspective is related to stability analysis through its focus on state evolution, DMP does not explicitly estimate Lyapunov exponents \citep{gallicchio2018local} or exact dynamical invariants \citep{manjunath2013echo}. Its purpose is practical one-shot reservoir compression that accounts jointly for reservoir structure and input-driven temporal behavior.


\section{Echo State Networks}
\label{sec:esn}

ESNs consist of three main components: an input layer, a recurrent reservoir, and a readout layer, as illustrated in Figure~\ref{fig1:esn_architecture}. The input is projected through a fixed matrix \(W^{in}\), while the reservoir is defined by a sparse recurrent matrix \(W\). Only the readout weights \(W^{out}\) are trained, enabling efficient learning without backpropagation through time.

The reservoir dynamics are given by
\begin{equation}
\begin{aligned}
\mathbf{x}(t+1) =\;& (1-a)\mathbf{x}(t) 
+ a\, f\!\big(W\mathbf{x}(t) + W^{\mathrm{in}}\mathbf{u}(t+1) 
+ W^{\mathrm{fb}}\mathbf{y}(t)\big),
\end{aligned}
\label{eq:esn_state}
\end{equation}
where \(\mathbf{x}(t) \in \mathbb{R}^{N}\) denotes the reservoir state, \(\mathbf{u}(t)\) the input, and \(\mathbf{y}(t)\) the output. We use \(t\) as the time index and \(N\) as the number of reservoir neurons to avoid notational ambiguity. The activation function \(f(\cdot)\) is typically a hyperbolic tangent, and the leak rate \(a\) controls the trade-off between short-term responsiveness and memory retention.

To ensure stable dynamics, the spectral radius is constrained as \(\rho(W) < 1\), which is a standard practical condition associated with the Echo State Property (ESP) \citep{jaeger2007optimization}. Although the exact echo-state regime also depends on factors such as leak rate, input scaling, and nonlinearity, spectral-radius control remains a widely used mechanism for maintaining stable input-driven dynamics.

The output is computed as
\begin{equation}
\mathbf{y}(t+1) = f^{\mathrm{out}}\left(W^{\mathrm{out}}\mathbf{x}(t+1)\right),
\label{eq:esn_output}
\end{equation}
where \(f^{\mathrm{out}}(\cdot)\) is typically the identity function for regression tasks.



\subsection{Dynamical Mode Pruning}
\label{sec:dmp}

\begin{figure*}[t]
\centering
\includegraphics[
width=0.99\linewidth,
height=0.99\linewidth,
keepaspectratio
]{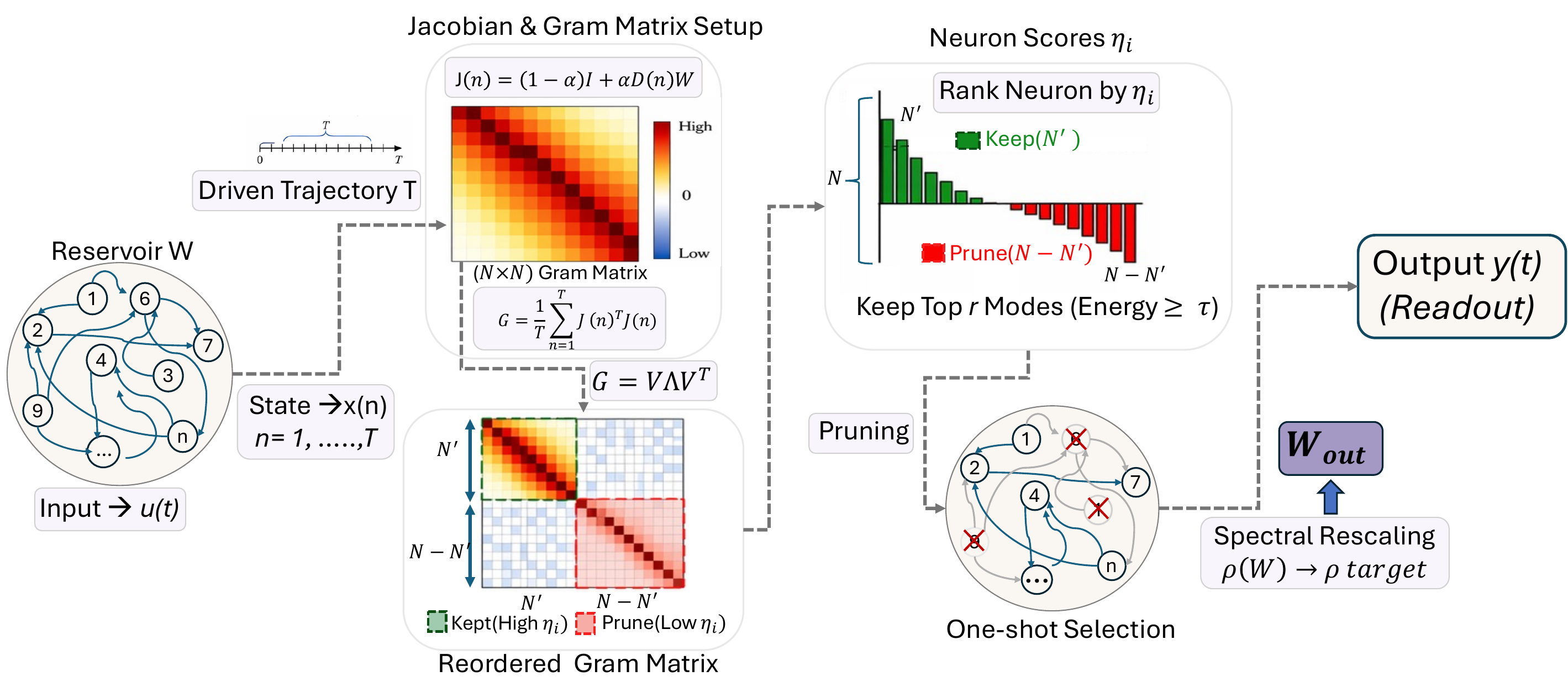}
\caption{
DMP framework. Reservoir states are collected from the driven ESN, and
trajectory-dependent Jacobians are aggregated into a Gramian capturing
the dominant dynamical modes. Neurons are ranked by their contribution to
these modes and pruned in a one-shot manner. Spectral-radius rescaling is
evaluated as an optional post-pruning step.
}
\label{fig:pruning_framework}
\end{figure*}

The overall pruning framework is illustrated in Figure~\ref{fig:pruning_framework}. We construct the pruning criterion from the local dynamics of the input-driven reservoir. In particular, the sensitivity of the state transition along the observed trajectory is first characterized through its Jacobian. These local sensitivities are then aggregated over time to obtain a global representation of the dynamical directions that are most strongly expressed during reservoir operation. Neuron importance is subsequently defined from the contribution of each state coordinate to these dominant directions.

For the ESN introduced in Section~\ref{sec:esn}, the reservoir state \(\mathbf{x}(t)\in\mathbb{R}^{N}\) evolves according to Equation~\eqref{eq:esn_state}. Around a particular state on the driven trajectory, a small perturbation of \(\mathbf{x}(t)\) induces a corresponding perturbation in \(\mathbf{x}(t+1)\). To first order, this local dependence can be expressed by the transition Jacobian 

\begin{equation} 
J(t) = \frac{\partial \mathbf{x}(t+1)} 
{\partial \mathbf{x}(t)}. 
\label{eq:jacobian_def} 
\end{equation}

For a leaky ESN with leak rate \(a\), differentiating the state-update equation with respect to the current reservoir state gives 

\begin{equation} 
J(t) = (1-a)I + aD(t)W, 
\label{eq:jacobian} 
\end{equation} 

where \(W\in\mathbb{R}^{N\times N}\) denotes the recurrent reservoir matrix and \(I\) is the identity matrix. The first term follows from the direct contribution of the previous state in the leaky update, whereas the second term represents the local recurrent response after modulation by the activation derivative.

To make this dependence explicit, we define the reservoir pre-activation as 

\begin{equation} 
\mathbf{z}(t) = W\mathbf{x}(t) + W^{\mathrm{in}}\mathbf{u}(t+1) + W^{\mathrm{fb}}\mathbf{y}(t), 
\label{eq:preactivation} 
\end{equation} 

from which the local activation sensitivity can be written as

\begin{equation} 
D(t) = \operatorname{diag} \left( f'\!\left(\mathbf{z}(t)\right) \right). 
\label{eq:Dt} 
\end{equation} 
Thus, the \(i\)-th diagonal element of \(D(t)\) is the derivative of the activation function evaluated at the pre-activation of reservoir neuron \(i\). Through Equation~\eqref{eq:jacobian}, this quantity modulates the effect of the recurrent connections on the local state transition.

The Jacobian in Equation~\eqref{eq:jacobian} is evaluated along the teacher-forced trajectory used for reservoir-state collection. Under this setting, \(\mathbf{y}(t)\) is supplied externally and is therefore treated as independent of \(\mathbf{x}(t)\) when computing the derivative. The expression in Equation~\eqref{eq:jacobian} consequently corresponds to the open-loop transition Jacobian. For an autonomous closed-loop ESN, where the output is itself a function of the reservoir state, the derivative would additionally contain the corresponding feedback chain-rule term involving \(W^{\mathrm{fb}}\) and \(W^{\mathrm{out}}\).

\begin{wrapfigure}{r}{0.54\linewidth}
\vspace{-10pt}
\centering
\begin{minipage}{0.98\linewidth}

\hrule
\vspace{2pt}
\makebox[\linewidth][c]{\textbf{DMP}}
\hrule
\vspace{4pt}

\footnotesize
\begin{algorithmic}[1]

\STATE \textbf{Input:}
reservoir \(W\in\mathbb{R}^{N\times N}\),
input matrix \(W^{\mathrm{in}}\),
feedback matrix \(W^{\mathrm{fb}}\)

\STATE \hspace{1.2em}
leak rate \(a\),
pre-activations
\(\{\mathbf{z}(t)\}_{t=1}^{T}\),
energy threshold \(\tau\)

\STATE \hspace{1.2em}
target size \(N'\),
optional target spectral radius
\(\rho_{\mathrm{target}}\)

\STATE \textbf{Output:}
pruned matrices
\(W_{\mathcal{I}}\),
\(W^{\mathrm{in}}_{\mathcal{I}}\), and
\(W^{\mathrm{fb}}_{\mathcal{I}}\)

\STATE \(\mathcal{G}\gets 0\)

\FOR{\(t=1,\ldots,T\)}

    \STATE
    \(D(t)\gets
    \operatorname{diag}
    \left(
    f'(\mathbf{z}(t))
    \right)\)

    \STATE
    \(J(t)\gets
    (1-a)I+aD(t)W\)

    \STATE
    \(\mathcal{G}\gets
    \mathcal{G}
    +
    \frac{1}{T}J(t)^{\top}J(t)\)

\ENDFOR

\STATE Compute
\(\mathcal{G}=V\Lambda V^{\top}\)

\STATE Sort eigenpairs such that
\(\lambda_1\geq\lambda_2\geq\cdots\geq\lambda_N\)

\STATE
\(\widetilde{\lambda}_k
\gets
\lambda_k/\sum_{j=1}^{N}\lambda_j\)

\STATE Find the smallest \(r\) satisfying
\(\sum_{k=1}^{r}\widetilde{\lambda}_k\geq\tau\)

\FOR{\(i=1,\ldots,N\)}

    \STATE
    \(\eta_i
    \gets
    \sum_{k=1}^{r}
    \widetilde{\lambda}_kV_{ik}^{2}\)

\ENDFOR

\STATE
\(\mathcal{I}
\gets
\operatorname{Top}\text{-}N'
(\boldsymbol{\eta})\)

\STATE
\(W_{\mathcal{I}}
\gets
W[\mathcal{I},\mathcal{I}]\)

\STATE
\(W^{\mathrm{in}}_{\mathcal{I}}
\gets
W^{\mathrm{in}}[\mathcal{I},:]\)

\STATE
\(W^{\mathrm{fb}}_{\mathcal{I}}
\gets
W^{\mathrm{fb}}[\mathcal{I},:]\)

\STATE \textbf{Optional:}
\(W_{\mathcal{I}}
\gets
\rho_{\mathrm{target}}
\dfrac{W_{\mathcal{I}}}
{\rho(W_{\mathcal{I}})}\)

\STATE \textbf{return}
\(W_{\mathcal{I}}\),
\(W^{\mathrm{in}}_{\mathcal{I}}\),
\(W^{\mathrm{fb}}_{\mathcal{I}}\)

\end{algorithmic}

\vspace{4pt}
\hrule

\end{minipage}
\vspace{-15pt}
\end{wrapfigure}

Equation~\eqref{eq:jacobian} characterizes the reservoir dynamics only in the neighborhood of the state visited at time \(t\). A pruning criterion based on a single \(J(t)\) would therefore depend on one local transition and would not represent the behavior of the reservoir over the full driven trajectory. We instead aggregate these local sensitivities over the post-washout state sequence and define the trajectory-averaged Jacobian Gramian as

\begin{equation} 
\mathcal{G} = \frac{1}{T} \sum_{t=1}^{T} J(t)^{\top}J(t), \label{eq:gramian} 
\end{equation} 

where \(T\) is the number of post-washout training time steps used for state collection.

Because each matrix \(J(t)^{\top}J(t)\) is symmetric and positive semidefinite, their average \(\mathcal{G}\in\mathbb{R}^{N\times N}\) has the same properties. For an arbitrary perturbation direction \(\boldsymbol{\delta}\in\mathbb{R}^{N}\), the quadratic form

\begin{equation} 
\boldsymbol{\delta}^{\top} 
\mathcal{G} 
\boldsymbol{\delta} 
= \frac{1}{T} \sum_{t=1}^{T} 
\left\| 
J(t)\boldsymbol{\delta} \right\|_{2}^{2} \label{eq:gramian_quadratic} 
\end{equation}

represents the time-averaged squared magnitude of the locally propagated perturbation. The eigendirections of \(\mathcal{G}\) can therefore be used to identify directions in reservoir-state space along which local perturbations are, on average, most strongly represented over the driven trajectory. Since the averaging is performed over time, transition patterns that occur persistently contribute repeatedly to \(\mathcal{G}\), whereas effects confined to a small number of time steps have proportionally smaller influence.

We express the Gramian through the eigendecomposition 

\begin{equation} 
\mathcal{G} = V\Lambda V^{\top}, \label{eq:gramian_eigendecomposition} 
\end{equation}

where \(\Lambda=\operatorname{diag}(\lambda_1,\ldots,\lambda_N)\) contains the non-negative eigenvalues and the columns of \(V\) contain the corresponding orthonormal eigenvectors. We order the eigenpairs such that

\begin{equation} 
\lambda_1 
\geq 
\lambda_2 
\geq 
\cdots 
\geq 
\lambda_N, 
\label{eq:eigenvalue_ordering} 
\end{equation}

so that eigenvectors associated with larger eigenvalues correspond to directions having larger average Jacobian energy along the trajectory.

To express the contribution of each mode relative to the total Gramian energy, we normalize the eigenvalues as 

\begin{equation} 
\widetilde{\lambda}_k 
= 
\frac{\lambda_k} 
{\sum_{j=1}^{N}\lambda_j}. 
\label{eq:eig_norm} 
\end{equation} 

The quantity \(\widetilde{\lambda}_k\) therefore gives the fraction of the total Gramian energy associated with mode \(k\). Rather than using all \(N\) eigendirections in the pruning criterion, we define a dominant subspace by retaining the smallest number \(r\) for which 

\begin{equation} 
\sum_{k=1}^{r} 
\widetilde{\lambda}_k 
\geq \tau, 
\label{eq:energy_rank} 
\end{equation} 

where \(\tau\in(0,1]\) is a prescribed energy threshold. Thus, \(r\) is determined directly by the spectrum of the trajectory-averaged Gramian: only as many leading modes as required to represent at least a fraction \(\tau\) of its total energy are retained.

This truncation defines the portion of the observed transition structure used for pruning. Directions associated with the remaining lower-energy modes do not enter the neuron score. Such directions may still encode infrequent or task-specific behavior; the truncation instead restricts the pruning criterion to the dominant transition structure represented by the chosen value of \(\tau\). Having defined the retained dynamical subspace, we next assign its modal energy to the individual reservoir coordinates. Let \(V_{ik}\) denote the \(i\)-th component of eigenvector \(k\). We define the importance of reservoir neuron \(i\) as 

\begin{equation} 
\eta_i = \sum_{k=1}^{r} 
\widetilde{\lambda}_k V_{ik}^{2}. 
\label{eq:importance} 
\end{equation} 

For a given mode \(k\), the squared component \(V_{ik}^{2}\) represents the participation of reservoir coordinate \(i\) in that eigendirection, while \(\widetilde{\lambda}_k\) weights this participation by the relative energy of the mode. Summing over the retained modes therefore associates each neuron with the portion of the dominant Gramian energy represented along its coordinate. The squared eigenvector components also make the score independent of the arbitrary sign of individual eigenvectors and ensure that \(\eta_i\) is non-negative. The resulting scores provide an ordering of the reservoir neurons. 
For a desired reduced reservoir size \(N'\), we define the retained index set as 

\begin{equation} 
\mathcal{I} = \operatorname{Top}\text{-}N' 
\left( \boldsymbol{\eta} \right),
\label{eq:retained_indices} 
\end{equation} 

where \(\boldsymbol{\eta} = (\eta_1,\ldots,\eta_N)^{\top}\). 
The reduced recurrent matrix can then be expressed as 

\begin{equation} 
W_{\mathcal{I}} = W[\mathcal{I},\mathcal{I}], 
\label{eq:pruned_W} 
\end{equation} 

with the corresponding input and feedback matrices given by 

\begin{equation} 
W^{\mathrm{in}}_{\mathcal{I}}
= W^{\mathrm{in}}[\mathcal{I},:] 
\label{eq:pruned_Win} 
\end{equation} 

and 

\begin{equation} 
W^{\mathrm{fb}}_{\mathcal{I}} 
= W^{\mathrm{fb}}[\mathcal{I},:]. 
\label{eq:pruned_Wfb} 
\end{equation} 

Restricting both rows and columns of \(W\) to the same index set preserves the recurrent interactions among the retained neurons, while applying the corresponding row indices to \(W^{\mathrm{in}}\) and \(W^{\mathrm{fb}}\) maintains dimensional consistency of the reduced ESN. The construction above defines a one-shot pruning procedure. The trajectory, Jacobians, Gramian, eigendecomposition, and neuron scores are computed once using the original reservoir, after which the selected index set \(\mathcal{I}\) is fixed. These operations are performed offline and do not contribute to inference-time computation. In the primary DMP configuration, the recurrent weights retained in the principal submatrix are left unchanged and the linear readout is refitted using the reduced reservoir states. To separately examine the effect of restoring the recurrent spectral scale after pruning, we additionally consider DMP+\(\rho\), in which the reduced recurrent matrix is rescaled according to \begin{equation} W_{\mathcal{I}} \leftarrow \rho_{\mathrm{target}} \frac{W_{\mathcal{I}}} {\rho(W_{\mathcal{I}})}, \label{eq:spectral_rescaling} \end{equation} where \(\rho(\cdot)\) denotes the spectral radius and \(\rho_{\mathrm{target}}\) is the prescribed target value. In either configuration, inference is subsequently performed using only the reduced reservoir and its refitted output layer.

\section{Experimental Design and Benchmark Analysis}
We evaluate DMP on time-series forecasting tasks across reservoir sizes, forecasting horizons, and pruning ratios. Comparisons include the unpruned ESN, representative ESN variants, and graph-based pruning baselines.

\paragraph{Benchmarks:}
We consider one synthetic benchmark and four real-world datasets. For the synthetic setting, we use the Mackey--Glass system \citep{mackey1977oscillation}, a standard benchmark for chaotic time-series prediction. Due to its delayed nonlinear dynamics, it is well suited for evaluating both long-term forecasting accuracy and memory capacity. The system is defined as
\begin{equation}
\frac{d o(t)}{d t}
=
\frac{0.2\, o(t-\alpha)}{1 + o(t-\alpha)^{10}}
-
0.1\, o(t),
\end{equation}
where \(\alpha\) is the delay parameter, set to \(\alpha = 30\) following \citep{jaeger2001echo}.

To complement this controlled setting, we further consider four real-world univariate time-series datasets: electricity demand from the Australian National Electricity Market \citep{australia_energy}, temperature data from the Australian Bureau of Meteorology \citep{bom_temperature}, wind data from the National Renewable Energy Laboratory (NREL) Wind Toolkit \citep{nrel_wind_toolkit}, and solar radiation data from the National Solar Radiation Database (NSRDB) \citep{nsrdb}. All datasets used in this study are public research or government datasets, and the original data providers are cited in the references. No private or personally identifiable data are used. 
We focus on univariate forecasting to isolate the effect of the pruning criterion without introducing additional confounders from multivariate feature interactions or task-specific output structures.

\paragraph{Data split and forecasting setup:}
All datasets are treated as univariate forecasting problems. We use a fixed split of 10\% initial washout, 70\% training, and 20\% testing, and evaluate forecasting at horizons \(h \in \{10,20,30\}\). Performance is measured using mean squared error (MSE). The additional
matched comparison reports normalized root mean squared error (NRMSE),
defined at forecasting horizon $h$ as
\begin{equation}
\operatorname{RMSE}_{h}
=
\sqrt{
\frac{1}{n_h}
\sum_{j=1}^{n_h}
\left(
\widehat{y}_{j}^{(h)}-y_{j}^{(h)}
\right)^2
},
\qquad
\operatorname{NRMSE}_{h}
=
\frac{\operatorname{RMSE}_{h}}
{\sigma\!\left(\mathbf{y}^{(h)}\right)},
\label{eq:nrmse}
\end{equation}
where $n_h$ is the number of test predictions at horizon $h$ and $\sigma(\mathbf{y}^{(h)})$ is the population standard deviation of the corresponding test targets. Each time series is linearly scaled to $[-1,1]$ using the minimum and maximum values from the washout and training portion only; the test data do not influence normalization.

The pruning ratios of 10\%, 20\%, and 30\% are reported as a sensitivity analysis rather than as test-set-based model selection. Unless otherwise stated, 20\% pruning is used as a fixed moderate-pruning setting chosen a priori to evaluate whether redundancy can be removed without strongly disrupting the reservoir dynamics. For Mackey--Glass, we generate 100{,}000 samples and follow the same protocol to ensure consistency.

\paragraph{Reservoir configurations:}
We examine how behavior varies with reservoir size. Both the standard ESN and DMP are evaluated using \(N \in \{100, 200, 300, 500, 700, 1000\}\). For baseline comparisons, we report results at \(N=1000\), which provides sufficient capacity for reliable estimation of dynamical influence and enables consistent comparison across datasets. Smaller reservoirs are included to study scaling behavior, where limited capacity may make neuron-importance estimates less stable and reduce the margin for safe pruning.

The reservoir is initialized with leak rate \(a=0.5\), and the readout uses ridge coefficient \(\beta=10^{-4}\). The energy threshold for mode selection is fixed at \(\tau=0.9\). As illustrated by the cumulative Gramian energy curve in Figure~\ref{fig:3}(b), \(\tau\) can be chosen using an elbow heuristic: we retain modes until the marginal increase in explained energy becomes small. Across datasets, this transition generally falls within \([0.85,0.95]\), supporting \(\tau=0.9\) as a fixed setting rather than a task-specific tuning parameter.


The benchmark, scaling, horizon, and pruning-ratio experiments are repeated over five random seeds to account for variability in initialization. The additional matched comparison is evaluated over ten paired random seeds.

Within each evaluation protocol, we keep the dataset split, forecasting horizons, evaluation metric, and number of random seeds fixed across all compared models. The main pruning comparison is between DMP and graph-centrality pruning baselines under the same target reservoir size and pruning ratio. We note, however, that architectural variants such as Leaky ESN and Deep ESN follow their standard configurations and should be interpreted as reference models rather than strictly controlled ablations. A fully matched hyperparameter comparison across all ESN variants is left for future work.

\paragraph{Controlled matched evaluation protocol:}
To provide a controlled evaluation of the pruning criterion, we conduct an additional matched comparison using an initial reservoir of $N=1000$ neurons and a retained reservoir of $N_{\mathrm{ret}}=800$, corresponding to 20\% neuron pruning. The comparison includes the full ESN, a freshly initialized smaller ESN with $N=800$, random pruning with and without spectral-radius rescaling, and DMP with and without spectral-radius rescaling. DMP without rescaling is treated as the main method, while DMP+$\rho$ is included as an ablation.

For each dataset, the methods are evaluated over ten paired random seeds. For a given seed, DMP and random pruning operate on the same original reservoir, use the same data split, and retain the same number of neurons. After pruning, the recurrent weights remain fixed and the linear readout is refitted using ridge regression. We report forecasting NRMSE together with the number of recurrent parameters, model size, inference latency, and speedup of the reduced reservoir. Offline pruning time is reported separately because it is incurred only once and is not part of final inference.

\begin{table}[!t]
\centering
\caption{
Illustrative benchmark comparison across five datasets at reservoir size
$N=1000$, forecasting horizon $h=20$, pruning ratio 20\%, and one
representative random seed. These results provide a detailed comparison with
the original ESN variants and graph-based pruning baselines; aggregate
five-seed forecasting results are reported separately in
Table~\ref{tab:aggregate_improvement_combined}. Runtime and memory are end-to-end offline measurements for each method. For
DMP, these measurements include the original dense Jacobian--Gramian
construction and eigendecomposition pipeline, and memory denotes peak
offline usage. They are therefore not directly comparable to the isolated
streamlined pruning time reported in
Table~\ref{tab:matched_pruning_efficiency}.
}
\label{tab:benchmark}

\small
\setlength{\tabcolsep}{5pt}        
\renewcommand{\arraystretch}{1.16} 

\begin{tabular}{llccc}
\toprule
\textbf{Dataset} & \textbf{Model} & \textbf{MSE $\downarrow$} & \textbf{Runtime (s) $\downarrow$} & \textbf{Peak Offline Memory (MB) $\downarrow$} \\

\multirow{6}{*}{\textbf{Mackey--Glass}}
& Base     & 0.018844 & 16.766 & 301.78 \\
& Leaky    & 0.958887 & 16.741 & 305.50 \\
& Deep     & 0.016490 & 42.897 & 1810.03 \\
& Betweenness & 0.015915 & 305.797 & 246.55 \\
& Closeness   & 0.029860 & 517.687 & 237.84 \\
& DMP     & \textbf{0.0101305} & 570.096 & 1836.86 \\

\midrule
\multirow{6}{*}{\textbf{Electricity}}
& Base     & 0.006728 & 5.663 & 123.53 \\
& Leaky    & 0.006941 & 5.659 & 126.73 \\
& Deep     & 0.006051 & 14.437 & 763.91 \\
& Betweenness & 0.029764 & 291.922 & 98.35 \\
& Closeness   & 0.005905 & 516.705 & 98.34 \\
& DMP     & \textbf{0.005762} & 371.940 & 1081.08 \\

\midrule
\multirow{6}{*}{\textbf{Temperature}}
& Base     & 0.159485 & 1.471 & 115.35 \\
& Leaky    & 0.069661 & 1.484 & 119.60 \\
& Deep     & 0.754666 & 3.743 & 714.42 \\
& Betweenness & 0.075101 & 295.583 & 91.54 \\
& Closeness   & 0.122108 & 512.516 & 91.54 \\
& DMP     & \textbf{0.0661202} & 495.039 & 725.13 \\

\midrule
\multirow{6}{*}{\textbf{Wind}}
& Base     & 0.034031 & 5.297 & 51.87 \\
& Leaky    & 0.033207 & 5.390 & 55.72 \\
& Deep     & 0.044787 & 13.682 & 341.21 \\
& Betweenness & 0.032941 & 288.588 & 39.98 \\
& Closeness   & 0.043228 & 511.121 & 39.98 \\
& DMP     & \textbf{0.031385} & 331.043 & 412.37 \\

\midrule
\multirow{6}{*}{\textbf{Solar}}
& Base     & 0.094400 & 19.640 & 362.88 \\
& Leaky    & 0.073327 & 20.255 & 363.20 \\
& Deep     & 0.430456 & 50.288 & 1850.92 \\
& Betweenness & 12.104737 & 313.336 & 297.12 \\
& Closeness   & 0.038933 & 518.997 & 297.12 \\
& DMP     & \textbf{0.010819} & 412.637 & 1601.42 \\

\bottomrule
\end{tabular}
\end{table}

\paragraph{Pruning protocol:}
We perform pruning in a one-shot manner. After training the initial ESN and collecting the driven trajectory, we compute a trajectory-averaged Jacobian Gramian and derive neuron importance scores from dominant modes. Neurons with the lowest scores are removed according to target pruning ratios of 10\%, 20\%, and 30\%, maintaining a simple procedure and confining additional cost to a single offline step.

The recurrent matrix is then reduced to the corresponding principal submatrix, with input and feedback matrices pruned accordingly to maintain dimensional consistency. The main DMP configuration uses the reduced recurrent matrix without
post-pruning spectral-radius rescaling. Spectral rescaling is applied only to the variants for which it is explicitly indicated. The readout layer is then retrained using closed-form ridge regression with regularization parameter \(\beta=10^{-4}\), while recurrent weights remain fixed. All experiments were implemented in Python. The original benchmark
experiments were executed on an NVIDIA RTX 5000 Ada GPU, while the matched latency measurements were collected on CPU using double-precision
inference.

\paragraph{Baselines and comparison models:}
We compare DMP with the original unpruned ESN, centrality-based pruning baselines based on betweenness and closeness, and architectural ESN variants including Leaky ESN and Deep ESN. The controlled comparison further includes a freshly initialized smaller ESN and matched random pruning. Additional matched comparisons with magnitude, activation-variance, Jacobian column-energy, and betweenness pruning are reported in the supplementary material.


\section{Results}
\paragraph{Benchmark comparison:}
Table~\ref{tab:benchmark} summarizes representative benchmark behavior across
five datasets at $N=1000$, while
Table~\ref{tab:aggregate_improvement_combined} reports the aggregate
five-seed MSE statistics. DMP achieves lower mean MSE than the unpruned ESN
across all datasets in the main evaluation configuration, although the
magnitude of improvement varies substantially across datasets and random
initializations. The representative comparison with structural baselines further suggests
that trajectory-dependent dynamical influence provides a useful signal for
reservoir refinement beyond static connectivity alone.

\begin{wrapfigure}{r}{0.45\linewidth}
\vspace{-1pt}
\centering

\hspace*{-0.03\linewidth}
\includegraphics[width=0.99\linewidth]{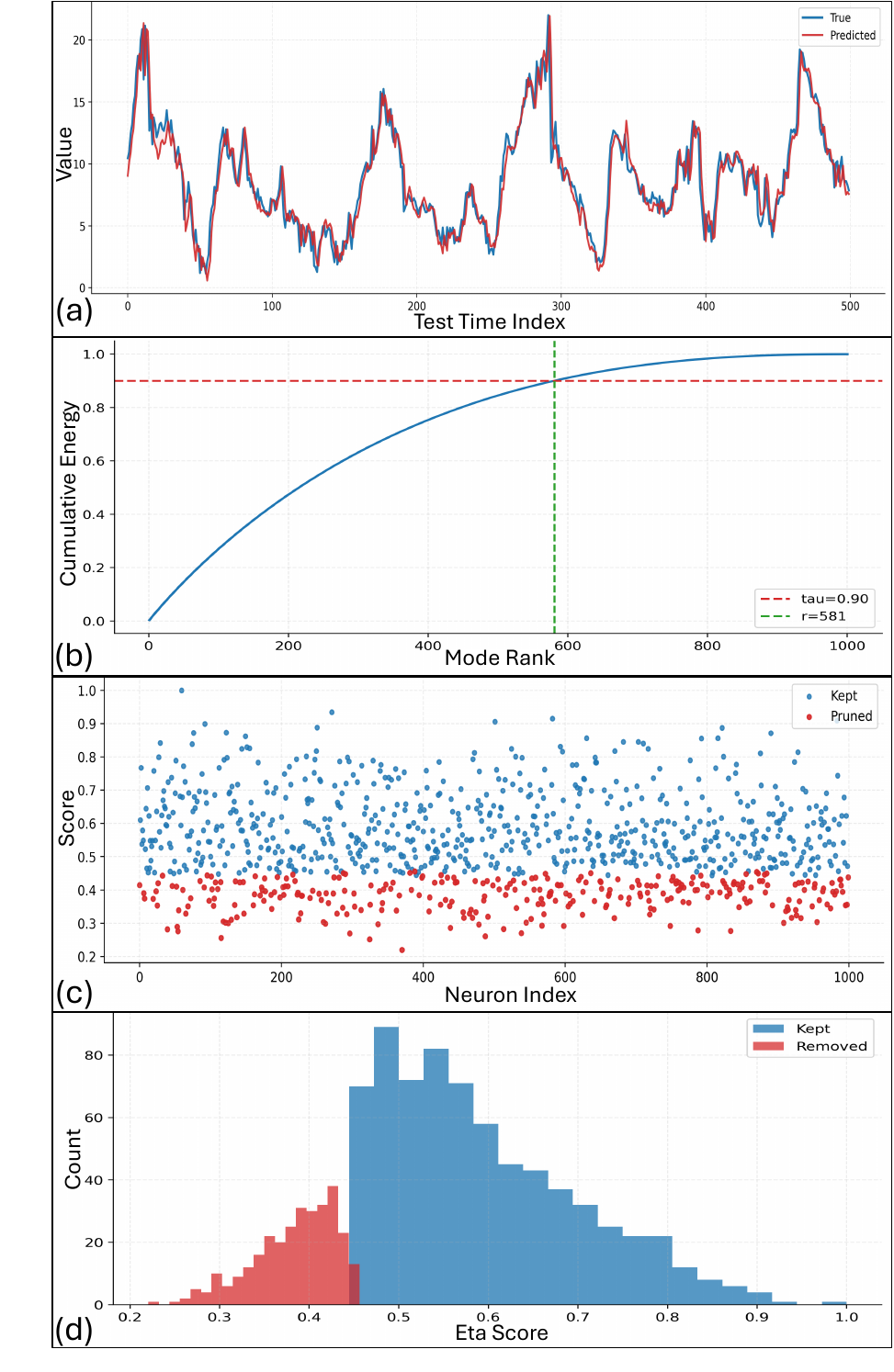}
\caption{
DMP analysis:
(a) Ground-truth and predicted time series after pruning.
(b) Cumulative energy of the trajectory-averaged Jacobian Gramian($\tau=0.9$) and the corresponding
mode rank indicated.
(c) DMP neuron-importance scores, distinguishing retained and pruned
reservoir neurons.
(d) Distribution of importance scores for the retained and pruned
neuron groups.
}
\label{fig:3}

\vspace{1pt}

\hspace*{0.03\linewidth}
\includegraphics[width=0.98\linewidth]{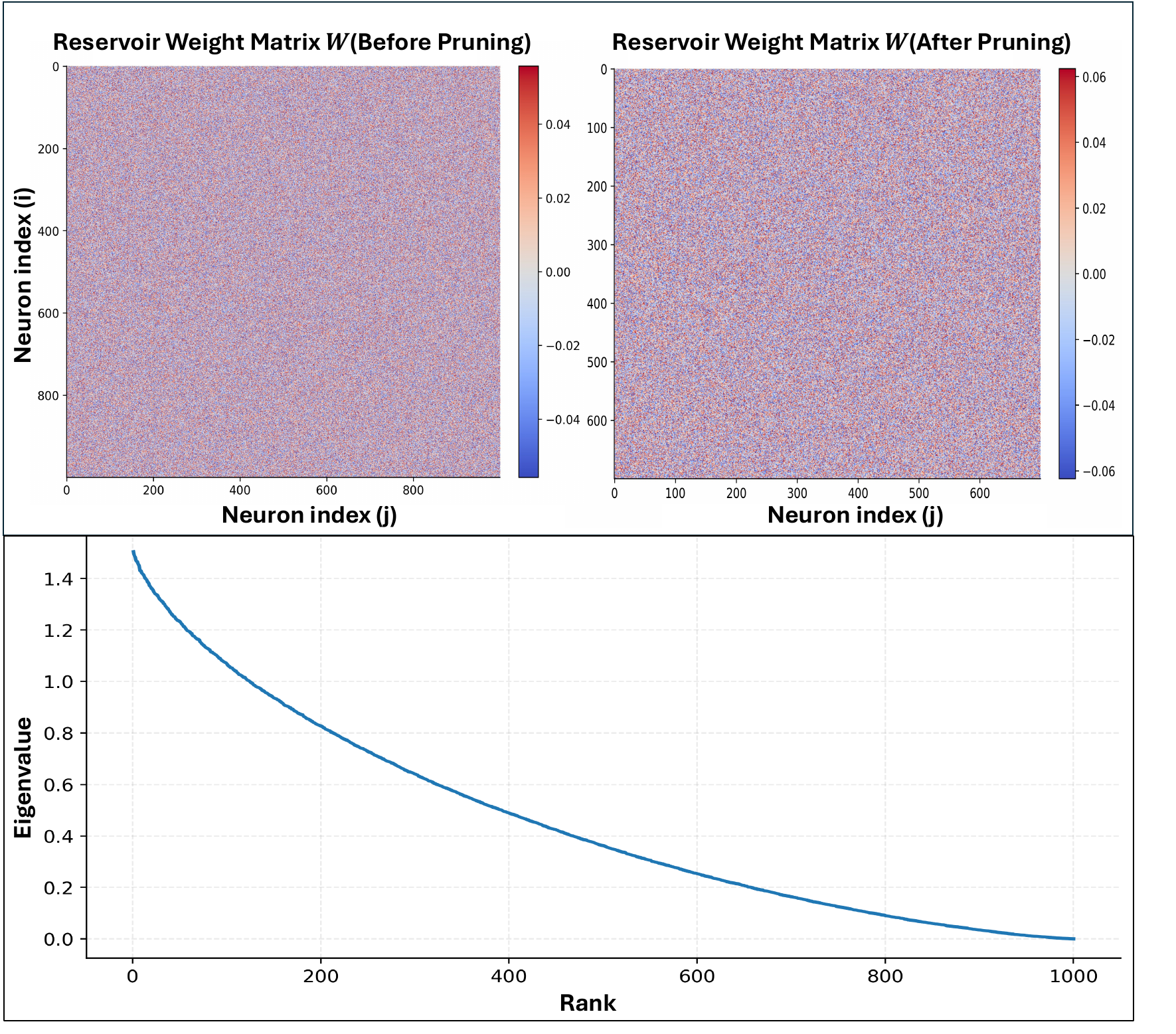}
\caption{
Representative structural analysis of DMP. Top: recurrent reservoir weight matrices before and after pruning. Bottom: eigenvalue spectrum of the trajectory-averaged Jacobian Gramian used to identify the dominant dynamical subspace.
}
\label{fig:weights}

\vspace{-50pt}
\end{wrapfigure}

The representative run in Table~\ref{tab:benchmark} illustrates that DMP can
produce a substantial reduction in forecasting error for a particular
reservoir initialization. However, the aggregate results provide a more
reliable assessment of performance. For Mackey--Glass, the five-seed mean MSE
decreases from 0.0323 to 0.0298, corresponding to an aggregate improvement of
approximately 7.7\%. The difference between the representative and aggregate
results reflects the variability of chaotic prediction across reservoir
initializations. Figure~\ref{fig:3} illustrates the pruning behavior: the
cumulative Gramian energy in (b) shows that much of the transition energy is
concentrated in a reduced set of modes, while (c) and (d) show a separation
between retained and pruned neurons. This supports the view that moderate
pruning can improve or preserve accuracy without disrupting the main
reservoir dynamics.

\paragraph{Scaling and horizon sensitivity:}

\begin{table}[t]
\centering
\caption{
Aggregate forecasting performance over five random seeds. MSE values are reported as mean $\pm$ standard deviation with 95\% confidence intervals in brackets. Negative lower bounds are truncated at zero because MSE is non-negative. Relative improvement is computed from the mean MSE of the Base ESN and DMP. Lower MSE and higher improvement are better.
}
\label{tab:aggregate_improvement_combined}

\small
\setlength{\tabcolsep}{3pt}
\renewcommand{\arraystretch}{1.10}

\begin{tabular}{@{}lccc@{}}
\toprule
\textbf{Dataset}
& \textbf{Base ESN}
& \textbf{DMP}
& \shortstack{\textbf{Mean MSE}\\\textbf{Reduction (\%)}} \\
\midrule

Mackey--Glass
& $0.0323 \pm 0.0190$ {\scriptsize [0.0087, 0.0559]}
& $\mathbf{0.0298 \pm 0.0142}$ {\scriptsize [0.0122, 0.0474]}
& $7.7$ \\

Electricity
& $0.0096 \pm 0.0060$ {\scriptsize [0.0021, 0.0171]}
& $\mathbf{0.0078 \pm 0.0045}$ {\scriptsize [0.0022, 0.0134]}
& $18.8$ \\

Temperature
& $0.0707 \pm 0.0580$ {\scriptsize [0.0000, 0.1427]}
& $\mathbf{0.0682 \pm 0.0415}$ {\scriptsize [0.0167, 0.1197]}
& $3.5$ \\

Wind
& $19.8013 \pm 42.6936$ {\scriptsize [0.0000, 72.8170]}
& $\mathbf{2.0186 \pm 4.2726}$ {\scriptsize [0.0000, 7.3249]}
& $89.8$ \\

Solar
& $1.3337 \pm 2.5818$ {\scriptsize [0.0000, 4.5396]}
& $\mathbf{1.2985 \pm 2.1542}$ {\scriptsize [0.0000, 3.9737]}
& $2.6$ \\

\bottomrule
\end{tabular}
\end{table}

The runtime and peak offline memory results in
Table~\ref{tab:benchmark} show that the original dense DMP implementation
incurs substantial offline overhead, mainly due to Jacobian--Gramian
construction and eigendecomposition. These measurements cover the broader
benchmark pipeline, whereas the offline time in
Table~\ref{tab:matched_pruning_efficiency} measures the isolated streamlined
one-shot pruning procedure. The two timing results therefore have different
computational scopes and should not be compared directly. In both cases, the
cost is incurred once during reservoir refinement and does not affect the memory footprint or inference latency of the final reduced ESN.

\begin{wraptable}{r}{0.58\linewidth}
\vspace{-10pt}
\centering
\begin{minipage}{\linewidth}
\caption{
Sensitivity and reliability analysis. Lower MSE is better; \(p\)-values are paired tests on log-transformed MSE across random seeds.
}
\label{tab:combined_wrap}

\footnotesize
\setlength{\tabcolsep}{4pt}
\renewcommand{\arraystretch}{1.05}

\textbf{Horizon Sensitivity}

\vspace{2pt}

\begin{tabular*}{\linewidth}{@{\extracolsep{\fill}}lccc}
\toprule
Dataset & 10 & 20 & 30 \\
\midrule
Mackey--Glass & 0.0365 & \textbf{0.0101} & 0.2505 \\
Electricity   & 0.0094 & \textbf{0.0058} & 0.0058 \\
Temperature   & \textbf{0.0429} & 0.0661 & 1.0612 \\
Wind          & 0.0480 & \textbf{0.0314} & 0.1142 \\
Solar         & 0.4317 & \textbf{0.0108} & 0.0728 \\
\bottomrule
\end{tabular*}

\vspace{6pt}

\scriptsize
\setlength{\tabcolsep}{2pt}

\textbf{Scaling Behavior}

\vspace{2pt}

\begin{tabular*}{\linewidth}{@{\extracolsep{\fill}}lccccc}
\toprule
Dataset & 100 & 200 & 300 & 500 & 700 \\
\midrule
Mackey--Glass (Base) & 0.0397 & \textbf{0.0300} & \textbf{0.0255} & 0.1416 & 1.8450 \\
Mackey--Glass (DMP)  & \textbf{0.0123} & 0.2506 & 0.4083 & \textbf{0.0317} & \textbf{0.0184} \\
Electricity (Base)   & 0.0147 & \textbf{0.0051} & 0.0068 & 0.0073 & 0.0060 \\
Electricity (DMP)    & \textbf{0.0051} & 0.0053 & \textbf{0.0051} & \textbf{0.0052} & \textbf{0.0054} \\
Temperature (Base)   & 0.0309 & 0.0316 & 0.0374 & 0.0502 & 0.0445 \\
Temperature (DMP)    & \textbf{0.0296} & \textbf{0.0303} & \textbf{0.0320} & \textbf{0.0298} & \textbf{0.0382} \\
Wind (Base)          & \textbf{0.0612} & 0.0570 & \textbf{0.0590} & 11.099 & \textbf{0.0860} \\
Wind (DMP)           & 0.1253 & \textbf{0.0533} & 0.0590 & \textbf{0.1299} & 0.1461 \\
Solar (Base)         & 2.4185 & \textbf{0.0306} & \textbf{0.0193} & 0.0353 & 0.2105 \\
Solar (DMP)          & \textbf{0.0207} & 0.0449 & 0.1395 & \textbf{0.0268} & \textbf{0.0622} \\
\bottomrule
\end{tabular*}

\vspace{6pt}

\footnotesize
\setlength{\tabcolsep}{3.5pt}
\renewcommand{\arraystretch}{1.08}

\textbf{Pruning Ratio Sensitivity and Significance}

\vspace{2pt}

\begin{tabular*}{\linewidth}{@{\extracolsep{\fill}}lcccc}
\toprule
Dataset & 10\% & 20\% & 30\% & \(p\) \\
\midrule
Mackey--Glass & 0.01298 & \textbf{0.01013} & 0.03542 & 0.072 \\
Electricity   & 0.00621 & \textbf{0.00576} & 0.00895 & \textbf{0.008} \\
Temperature   & 0.07085 & \textbf{0.06612} & 0.15123 & \textbf{0.036} \\
Wind          & 0.03342 & \textbf{0.03139} & 0.07261 & \textbf{0.014} \\
Solar         & 0.01894 & \textbf{0.01082} & 0.09452 & \textbf{0.028} \\
\bottomrule
\end{tabular*}

\end{minipage}

\vspace{-20pt}
\end{wraptable}

Table~\ref{tab:combined_wrap} shows that DMP is capacity- and horizon-dependent. Smaller reservoirs leave less redundancy to remove, while larger reservoirs provide a richer state space in which dominant and redundant components can be separated more clearly. The trend is not monotonic because changing \(N\) also changes the random state space and Gramian conditioning. Across horizons, DMP performs strongly at \(h=20\) and remains competitive in several shorter- and longer-horizon settings, although longer horizons can degrade performance, as seen for Temperature at \(h=30\).

\paragraph{Pruning ratio analysis and reliability:}
Table~\ref{tab:combined_wrap} reports pruning-ratio sensitivity and paired statistical tests at \(h=20\). Moderate pruning at 20\% gives the strongest results within this sensitivity analysis, suggesting that DMP removes redundancy while preserving the core transition structure. With 10\% pruning, the gains are smaller because limited redundancy is removed, while 30\% pruning often degrades performance, especially on Temperature and Wind, indicating that aggressive pruning can remove components needed for stable signal propagation. The \(p\)-values provide supporting evidence for improvement on most real-world datasets, but because only five seeds are available and MSE values can be skewed, they should not be interpreted as definitive distributional claims. Mackey--Glass has a higher \(p\)-value, consistent with the higher variance expected in chaotic prediction. Table~\ref{tab:aggregate_improvement_combined} shows that DMP lowers the aggregate mean error, although the confidence intervals remain wide for high-variance datasets such as Wind and Solar. Figure~\ref{fig:weights} further shows that the pruned reservoir retains dominant connectivity patterns among the selected units while removing

lower-ranked components.

\paragraph{Matched ablation and final-model efficiency:}

Table~\ref{tab:matched_pruning_efficiency} presents the controlled comparison for the reduction $N=1000\rightarrow800$, corresponding to 20\% neuron pruning, over ten paired random seeds. The comparison separates the effect of the DMP criterion from the general effect of reducing reservoir size by including the full ESN, a freshly initialized $N=800$ ESN, random pruning with and without spectral-radius rescaling, and DMP with and without spectral-radius rescaling.

The DMP score provides a dynamics-aware interpretation of neuron importance by measuring how strongly each neuron participates in the dominant trajectory-dependent transition modes of the reservoir. High-scoring neurons contribute more strongly to the state-space directions through which input-driven perturbations propagate, whereas low-scoring neurons contribute less to the dominant observed transition structure and are therefore candidates for removal. In the controlled matched evaluation, DMP without spectral-radius rescaling improves upon the full ESN on all five datasets and achieves the lowest mean NRMSE in each comparison. It also obtains lower mean NRMSE than matched random pruning across all datasets, with the clearest separation on Mackey--Glass and smaller margins on the real-world datasets, particularly Solar and Wind. These results suggest that reservoir-size reduction itself contributes to the observed gains, while DMP provides a principled, trajectory-dependent criterion for deciding which neurons should be retained rather than relying on an arbitrary pruning mask.

Reducing the reservoir from 1000 to 800 neurons decreases the number of recurrent parameters from $1{,}000{,}000$ to $640{,}000$, an exact reduction of 36\%. The measured model size decreases from 7.798 MB to 5.017 MB, corresponding to a 35.7\% reduction in model storage. Across the five datasets, inference latency decreases from approximately $123$--$129~\mu$s per step to $64$--$66~\mu$s per step, yielding a speedup between $1.90\times$ and $1.97\times$.

Spectral-radius rescaling does not improve the matched forecasting results. DMP without rescaling achieves lower mean NRMSE than DMP+$\rho$ across all five datasets. We therefore retain DMP without rescaling as the main configuration and report DMP+$\rho$ as an ablation. The DMP offline pruning time is approximately $0.13$ seconds and is incurred only once; it does not affect the inference-time efficiency of the final reduced reservoir.


\subsection{Limitations and challenges}
\label{sec:limitations}

DMP introduces offline cost from Jacobian aggregation and Gramian
eigendecomposition. A direct dense implementation of Algorithm~1 requires $\mathcal{O}(TN^3)$ time to form $J(t)^{\top}J(t)$ across $T$ time steps, plus $\mathcal{O}(N^3)$ time for the eigendecomposition of the resulting Gramian. This cost is incurred only once during pruning and does not affect the subsequent inference cost of the refined ESN, but it makes the method less suitable for online pruning without approximation. In practice, the cost can be reduced by exploiting the diagonal structure of $D(t)$, reservoir sparsity, randomized eigensolvers, truncated eigendecomposition, or low-rank updates.

DMP also depends on sufficient reservoir capacity. In very small reservoirs,
the baseline ESN may remain competitive because there is limited redundancy
to remove. Since DMP uses a trajectory-averaged summary, neurons that matter
only during rare transient events may be under-valued, motivating
transient-aware or windowed variants. The one-shot score is also an
approximation of the post-pruning dynamics: after neurons are removed, the
Gramian computed from the original reservoir no longer exactly describes the
reduced system. This mismatch becomes more important at aggressive pruning
ratios, as seen in the degradation at 30\%. In practice, a sharp drop in
held-out performance, a large change in spectral radius after pruning, or a
strong shift in the Gramian eigenvalue spectrum can indicate that the
one-shot approximation is becoming unreliable.

\begin{table*}[!t]
\centering
\caption{
Controlled matched pruning comparison and final-model efficiency at forecasting horizon $h=20$. The original reservoir contains $N=1000$ neurons and all reduced models contain $N_{\mathrm{ret}}=800$ neurons, corresponding to 20\% neuron pruning. Results are reported as mean $\pm$ standard deviation over ten paired random seeds. Panel (a) reports forecasting NRMSE, where lower is better. Random and DMP denote pruning without spectral-radius rescaling, whereas Random+$\rho$ and DMP+$\rho$ apply post-pruning rescaling. Panel (b) compares the inference efficiency of the full ESN and the main
DMP configuration. Offline DMP pruning time is incurred once and is not included in inference latency. Bold values indicate the lowest mean NRMSE for each dataset.
}
\label{tab:matched_pruning_efficiency}

\scriptsize
\setlength{\tabcolsep}{3.2pt}
\renewcommand{\arraystretch}{1.15}

\resizebox{\linewidth}{!}{%
\begin{tabular}{lcccccc}
\toprule
\multicolumn{7}{c}{\textbf{(a) Matched forecasting comparison}} \\
\midrule
\textbf{Dataset}
& \textbf{Full ESN}
& \textbf{Fresh $N=800$}
& \textbf{Random}
& \textbf{Random+$\rho$}
& \textbf{DMP}
& \textbf{DMP+$\rho$} \\
\midrule

Electricity
& $1.1074 \pm 0.1515$
& $1.0275 \pm 0.0731$
& $0.8880 \pm 0.0315$
& $1.0483 \pm 0.1542$
& $\mathbf{0.8854 \pm 0.0066}$
& $1.0893 \pm 0.1825$ \\

Mackey--Glass
& $0.4094 \pm 0.0278$
& $0.4132 \pm 0.0107$
& $0.4514 \pm 0.0239$
& $0.4016 \pm 0.0206$
& $\mathbf{0.4015 \pm 0.0259}$
& $0.4048 \pm 0.0279$ \\

Solar
& $0.8989 \pm 0.0921$
& $0.8591 \pm 0.0477$
& $0.7641 \pm 0.0116$
& $0.8407 \pm 0.0816$
& $\mathbf{0.7640 \pm 0.0110}$
& $0.8698 \pm 0.0486$ \\

Temperature
& $1.3084 \pm 0.1002$
& $1.2426 \pm 0.1302$
& $1.1423 \pm 0.0650$
& $1.2485 \pm 0.1419$
& $\mathbf{1.1321 \pm 0.0571}$
& $1.1885 \pm 0.1035$ \\

Wind
& $1.2115 \pm 0.0824$
& $1.2011 \pm 0.1412$
& $1.0661 \pm 0.0360$
& $1.2394 \pm 0.0806$
& $\mathbf{1.0660 \pm 0.0435}$
& $1.1968 \pm 0.1004$ \\

\midrule
\multicolumn{7}{c}{\textbf{(b) Full ESN versus main DMP efficiency}} \\
\midrule

\textbf{Dataset}
& \textbf{Full latency}
& \textbf{DMP latency}
& \textbf{Speedup}
& \textbf{Model size}
& \textbf{Recurrent parameters}
& \textbf{DMP offline time} \\

& \multicolumn{2}{c}{\textbf{($\mu$s/step)}}
& \textbf{($\times$)}
& \textbf{Full / DMP (MB)}
& \textbf{Full / DMP}
& \textbf{(s)} \\
\midrule

Electricity
& $123.41 \pm 4.68$
& $64.39 \pm 0.45$
& $1.92 \pm 0.08$
& $7.798 / 5.017$
& $1{,}000{,}000 / 640{,}000$
& $0.132 \pm 0.004$ \\

Mackey--Glass
& $129.37 \pm 5.43$
& $65.62 \pm 1.35$
& $1.97 \pm 0.09$
& $7.798 / 5.017$
& $1{,}000{,}000 / 640{,}000$
& $0.134 \pm 0.004$ \\

Solar
& $126.98 \pm 7.21$
& $65.15 \pm 0.23$
& $1.95 \pm 0.11$
& $7.798 / 5.017$
& $1{,}000{,}000 / 640{,}000$
& $0.130 \pm 0.001$ \\

Temperature
& $125.37 \pm 9.10$
& $64.61 \pm 0.24$
& $1.94 \pm 0.15$
& $7.798 / 5.017$
& $1{,}000{,}000 / 640{,}000$
& $0.131 \pm 0.002$ \\

Wind
& $122.95 \pm 5.79$
& $64.87 \pm 0.55$
& $1.90 \pm 0.08$
& $7.798 / 5.017$
& $1{,}000{,}000 / 640{,}000$
& $0.133 \pm 0.002$ \\

\bottomrule
\end{tabular}%
}
\end{table*}


Finally, this study focuses on univariate forecasting. Extensions to
multivariate forecasting, classification, and broader temporal
decision-making tasks remain important directions for future work.

\paragraph{Code availability:}
The implementation, experiment configurations, and example time-series datasets are available at \url{https://github.com/Laudarisd/dmp_esn/}. The repository includes scripts and configuration files for reproducing the reported experiments and allows DMP and the comparison methods to be applied to additional univariate time series using the documented CSV format.


\section{Conclusion}
This work introduces DMP as a dynamics-aware approach to reservoir pruning, where neurons are ranked according to their participation in dominant trajectory-dependent transition modes rather than through static connectivity or activation statistics alone. The results indicate that fixed random reservoirs can contain substantial dynamical redundancy and that removing low-contribution components can simplify the reservoir while preserving its useful predictive behavior. By linking neuron importance directly to the input-driven state-transition structure, DMP provides a more principled view of reservoir refinement and helps distinguish dynamically relevant components from those that contribute little to the observed trajectory. More broadly, the Jacobian--Gramian perspective offers a useful framework for understanding and designing reservoir systems beyond pruning, with potential extensions to adaptive reservoir construction, task-dependent capacity selection, transient- or regime-aware scoring, and scalable dynamical model reduction for larger recurrent systems.

\section*{Conflict of Interest Statement}

The author declares that there is no conflict of interest.

\bibliographystyle{plainnat}
\bibliography{references}


%
%

\clearpage
\appendix

\setcounter{table}{0}
\renewcommand{\thetable}{S\arabic{table}}
\setcounter{figure}{0}
\renewcommand{\thefigure}{S\arabic{figure}}

\section{Additional Experimental Results}
\label{app:additional_results}

This supplement provides analyses that complement the controlled ablation in
our main paper. Section~\ref{app:additional_baselines} compares DMP with
additional pruning criteria under a common protocol.
Section~\ref{app:paired_consistency} examines whether the observed differences
are consistent across paired random seeds.
Section~\ref{app:temperature_case} presents detailed accuracy and efficiency visualizations for Temperature, where several pruning methods perform similarly.

\subsection{Additional Pruning Criteria}
\label{app:additional_baselines}

The main paper isolates the effects of reservoir reduction, neuron selection,
and spectral-radius rescaling using the full ESN, a freshly initialized
smaller ESN, random pruning, and DMP. Table~\ref{tab:supp_matched_pruning}
extends this comparison to magnitude, activation-variance, Jacobian
column-energy, and betweenness pruning.

These criteria represent complementary views of neuron importance. Magnitude
pruning uses recurrent-weight strength, activation-variance pruning uses
variation in the observed reservoir states, Jacobian column-energy pruning
uses total transition sensitivity, and betweenness pruning uses structural
importance in the reservoir graph.

The Jacobian column-energy baseline ranks neuron \(i\) using
\[
s_i^{\mathrm{Jac}}
=
\mathcal{G}_{ii}
=
\frac{1}{T}\sum_{t=1}^{T}
\left\|J_{:,i}(t)\right\|_2^2.
\]
This score measures the total trajectory-averaged transition energy associated
with one reservoir-state direction. DMP instead ranks neurons by their
contribution to the retained dominant modes. The comparison therefore tests
whether dominant-mode selection provides useful information beyond the full
Gramian diagonal.

All reduced reservoirs contain \(800\) neurons and are obtained from an
initial reservoir of \(1000\) neurons. Every pruning method in
Table~\ref{tab:supp_matched_pruning} uses post-pruning spectral-radius
rescaling. This common setting isolates differences among the selection
criteria; it does not replace the unscaled DMP configuration used as the main
method in the paper.

No criterion performs best on every dataset. DMP+\(\rho\) gives the lowest
mean NRMSE on Temperature, while other methods lead on the remaining datasets.

\begin{table*}[!h]
\centering
\caption{
Expanded matched comparison of neuron-selection criteria at forecasting
horizon \(h=20\). Values are mean \(\pm\) standard deviation over ten paired
random seeds; lower NRMSE is better. All reduced reservoirs contain
\(800\) neurons, and all pruning methods in this table use post-pruning
spectral-radius rescaling.
}
\label{tab:supp_matched_pruning}
\small
\resizebox{\linewidth}{!}{%
\begin{tabular}{lcccccccc}
\toprule
\textbf{Dataset}
& \textbf{Full ESN}
& \textbf{Fresh smaller ESN}
& \textbf{Random+\(\rho\)}
& \textbf{Magnitude+\(\rho\)}
& \textbf{Activation variance+\(\rho\)}
& \textbf{Jacobian col. energy+\(\rho\)}
& \textbf{Betweenness+\(\rho\)}
& \textbf{DMP+\(\rho\)} \\
\midrule
Electricity
& $1.1074 \pm 0.1515$
& $\mathbf{1.0275 \pm 0.0731}$
& $1.0483 \pm 0.1542$
& $1.0349 \pm 0.1436$
& $1.1351 \pm 0.1697$
& $1.0677 \pm 0.1052$
& $1.0882 \pm 0.1609$
& $1.0893 \pm 0.1825$ \\

Mackey--Glass
& $0.4094 \pm 0.0278$
& $0.4132 \pm 0.0107$
& $0.4016 \pm 0.0206$
& $0.4044 \pm 0.0285$
& $\mathbf{0.3778 \pm 0.0252}$
& $0.4172 \pm 0.0165$
& $0.3950 \pm 0.0165$
& $0.4048 \pm 0.0279$ \\

Solar
& $0.8989 \pm 0.0921$
& $0.8591 \pm 0.0477$
& $\mathbf{0.8407 \pm 0.0816}$
& $0.8681 \pm 0.0654$
& $0.9536 \pm 0.0920$
& $0.8575 \pm 0.0649$
& $0.8654 \pm 0.0524$
& $0.8698 \pm 0.0486$ \\

Temperature
& $1.3084 \pm 0.1002$
& $1.2426 \pm 0.1302$
& $1.2485 \pm 0.1419$
& $1.1981 \pm 0.1207$
& $1.7147 \pm 0.5623$
& $1.2080 \pm 0.0923$
& $1.2054 \pm 0.0861$
& $\mathbf{1.1885 \pm 0.1035}$ \\

Wind
& $1.2115 \pm 0.0824$
& $1.2011 \pm 0.1412$
& $1.2394 \pm 0.0806$
& $\mathbf{1.1707 \pm 0.0828}$
& $1.4719 \pm 0.2237$
& $1.1850 \pm 0.0513$
& $1.2544 \pm 0.1398$
& $1.1968 \pm 0.1004$ \\
\bottomrule
\end{tabular}%
}
\end{table*}

\subsection{Paired-Seed Consistency}
\label{app:paired_consistency}

Table~\ref{tab:supp_matched_pruning} compares the average performance of
different neuron-selection criteria. To complement these aggregate results,
Table~\ref{tab:paired_consistency} examines the consistency of the two main
comparisons across the ten paired random seeds.

The first comparison evaluates DMP against matched random pruning using
\(\Delta_{\mathrm{D-R}}\), defined as the NRMSE of DMP minus that of random
pruning. Negative values therefore indicate lower error for DMP. The second
comparison evaluates the effect of post-pruning spectral-radius rescaling
using \(\Delta_{\rho}\), defined as the NRMSE of DMP+\(\rho\) minus that of
unscaled DMP. Positive values therefore indicate lower error without
spectral-radius rescaling.

\begin{table}[!h]
\centering
\caption{
Seed-level consistency at \(N=1000\rightarrow800\) and \(h=20\).
\(\Delta_{\mathrm{D-R}}\) denotes DMP NRMSE minus matched random-pruning
NRMSE, while \(\Delta_{\rho}\) denotes DMP+\(\rho\) NRMSE minus DMP NRMSE.
Negative \(\Delta_{\mathrm{D-R}}\) values favour DMP, whereas positive
\(\Delta_{\rho}\) values favour DMP without spectral-radius rescaling.
Win counts are calculated over ten paired random seeds.
}
\label{tab:paired_consistency}

\small
\setlength{\tabcolsep}{4.5pt}
\renewcommand{\arraystretch}{1.12}

\begin{tabular}{lrrrrr}
\toprule
\textbf{Dataset}
& \textbf{Mean $\Delta_{\mathrm{D-R}}$}
& \textbf{Median $\Delta_{\mathrm{D-R}}$}
& \textbf{DMP wins}
& \textbf{Mean $\Delta_{\rho}$}
& \textbf{No-rescale wins} \\
\midrule
Electricity
& $-0.0026$
& $-0.0015$
& $6/10$
& $ 0.2039$
& $10/10$ \\

Mackey--Glass
& $-0.0499$
& $-0.0460$
& $9/10$
& $ 0.0033$
& $6/10$ \\

Solar
& $-0.0001$
& $-0.0002$
& $5/10$
& $ 0.1058$
& $10/10$ \\

Temperature
& $-0.0102$
& $-0.0060$
& $6/10$
& $ 0.0564$
& $8/10$ \\

Wind
& $-0.0001$
& $-0.0003$
& $5/10$
& $ 0.1308$
& $9/10$ \\
\bottomrule
\end{tabular}
\end{table}

The paired results show that the advantage of DMP over random pruning is
dataset-dependent. The clearest separation occurs on Mackey--Glass, where
both the mean and median differences favour DMP achieves lower NRMSE in nine of ten paired runs. Electricity and Temperature also show favourable
mean and median differences, although the margins are smaller and DMP wins
six of ten runs in each case. On Solar and Wind, the mean differences are
close to zero and the win counts are evenly split, indicating that DMP and
matched random pruning perform similarly for these datasets.

The spectral-radius ablation shows a more consistent pattern. The mean
\(\Delta_{\rho}\) is positive for all five datasets, indicating lower mean
NRMSE for DMP without post-pruning rescaling. This effect is strongest on
Electricity, Solar, and Wind, where the unscaled configuration also wins in
nine or ten of the paired runs. Temperature shows the same tendency in eight
of ten runs. On Mackey--Glass, the mean difference is small and the
no-rescaling configuration wins six of ten runs, suggesting that the two
configurations perform similarly for this dataset. Overall, these results
support using unscaled DMP as the main configuration while treating
DMP+\(\rho\) as an ablation.

\subsection{Temperature Case Study}
\label{app:temperature_case}

Temperature is examined in greater detail because DMP+\(\rho\) achieves the
lowest mean error among the uniformly rescaled criteria, while its margin
over magnitude, betweenness, and Jacobian column-energy pruning remains
small. This case illustrates why average accuracy, between-seed variation,
and inference cost should be considered together.

\begin{figure*}[!h]
\centering
\includegraphics[width=0.90\textwidth]
{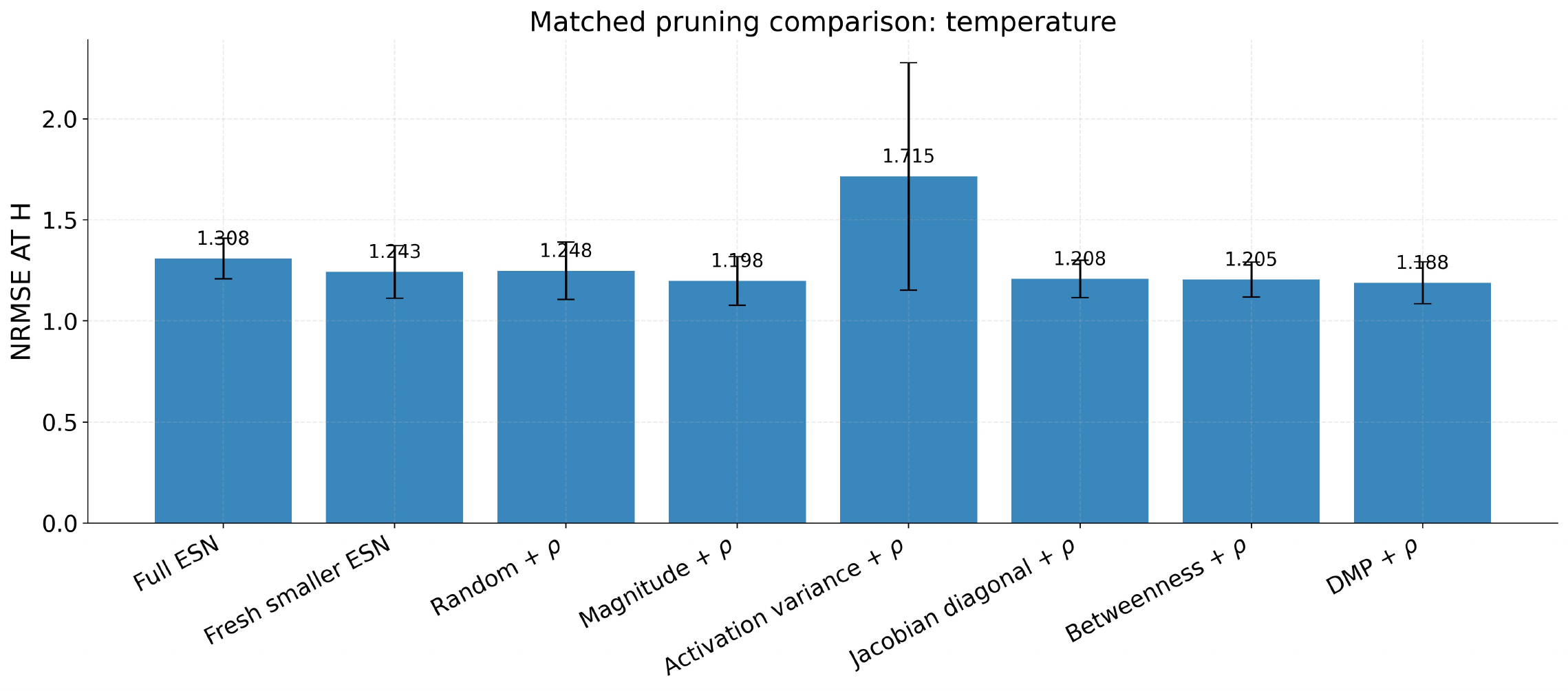}
\caption{
Visual comparison of the matched Temperature results in
Table~\ref{tab:supp_matched_pruning}. Bars show mean NRMSE at \(h=20\), and
error bars show one standard deviation over ten paired random seeds. All
pruning methods shown here use post-pruning spectral-radius rescaling.
DMP+\(\rho\) obtains the lowest mean error, although several criteria remain
close.
}
\label{fig:supp_temperature_comparison}
\end{figure*}

Figure~\ref{fig:supp_temperature_comparison} visualizes the same ten-seed
results reported in Table~\ref{tab:supp_matched_pruning}. It is a graphical
summary of the matched comparison rather than a separate experiment.

\begin{figure*}[!h]
\centering
\includegraphics[width=0.90\textwidth]
{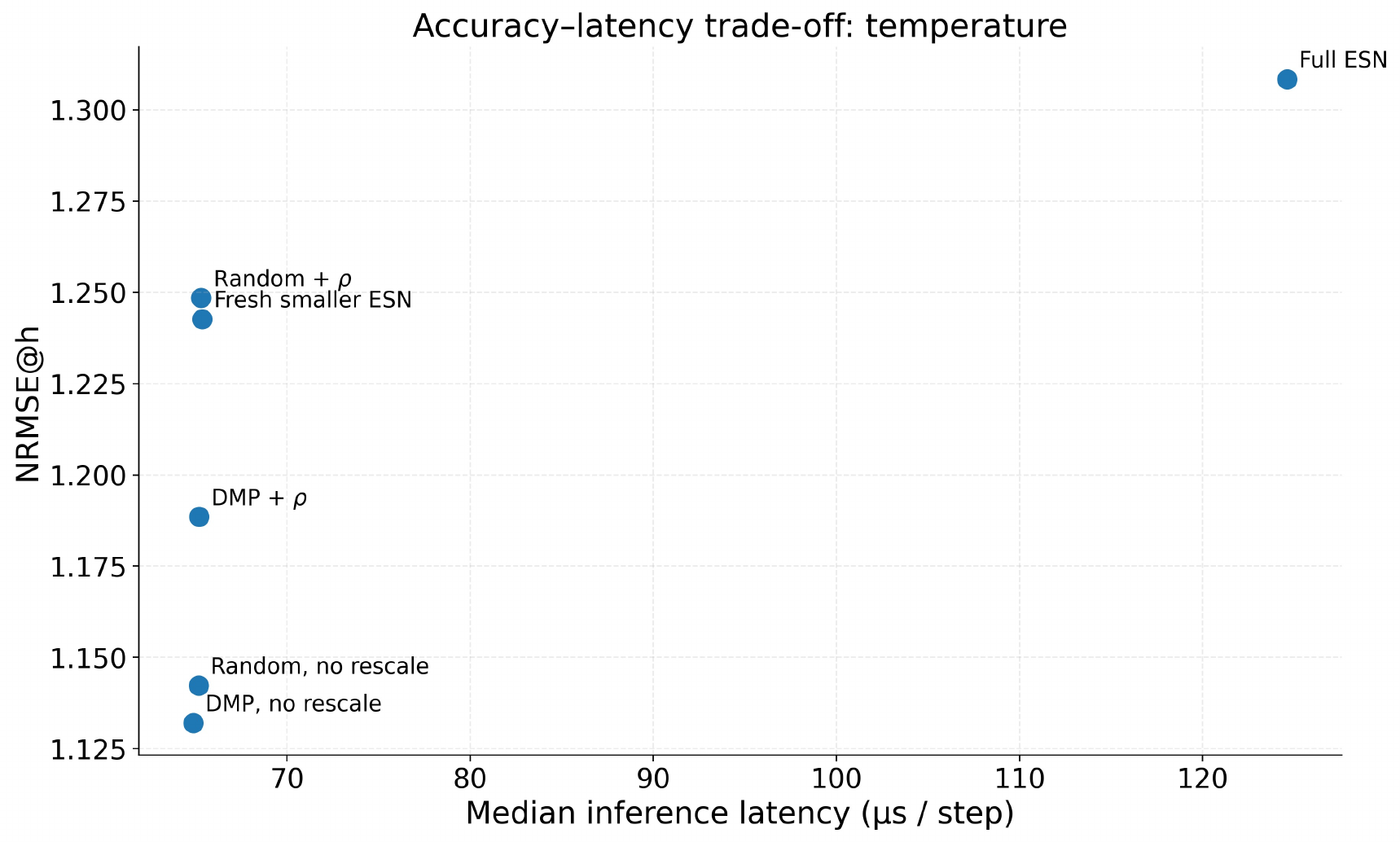}
\caption{
Accuracy--latency trade-off on Temperature after reducing the reservoir
from \(N=1000\) to \(N=800\). This figure complements the accuracy-only
comparison in Figure~\ref{fig:supp_temperature_comparison} by showing the
relationship between forecasting error and final-model inference cost.
Lower values on both axes are preferred.
}
\label{fig:supp_temperature_tradeoff}
\end{figure*}

Figure~\ref{fig:supp_temperature_tradeoff} provides a complementary view by
relating forecasting accuracy to final-model inference latency. Unlike
Figure~\ref{fig:supp_temperature_comparison}, which compares the uniformly
rescaled criteria, the accuracy--latency plot also includes the main DMP
configuration without rescaling. The reduced reservoirs occupy a similar
low-latency range, while unscaled DMP gives the lowest mean Temperature
NRMSE.






\end{document}